\documentclass[10pt,twocolumn,letterpaper]{article}

\usepackage[pagenumbers]{cvpr} 
\usepackage{cvpr}
\usepackage{times}
\usepackage{epsfig}
\usepackage{graphicx}
\usepackage{amsmath}
\usepackage{amssymb}
\usepackage{accents}
\usepackage{comment}
\usepackage{booktabs}
\usepackage{enumitem}
\usepackage{makecell,multirow,tabularx}
\usepackage{booktabs}
\usepackage{multirow}
\usepackage{colortbl}
\usepackage{multirow}
\usepackage[pagebackref=true,breaklinks=true,letterpaper=true,colorlinks,bookmarks=false]{hyperref}
\newcommand*{\Perm}[2]{{}^{#1}\!P_{#2}}%
\newcommand*{\Comb}[2]{{}^{#1}C_{#2}}%

\begin{document}

\title{Synthetic Forehead-creases Biometric Generation for Reliable User Verification}

\author{Abhishek Tandon\textsuperscript{1}\quad Geetanjali Sharma\textsuperscript{1}\quad Gaurav Jaswal \textsuperscript{2}\\Aditya Nigam \textsuperscript{1}\quad Raghavendra Ramachandra\textsuperscript{3}\\
\textsuperscript{1}Indian Institute of Technology  Mandi, India \\
\textsuperscript{2}Technology Innovation Hub - Indian Institute of Technology, Mandi\\
\textsuperscript{3}Norwegian University of Science and Technology (NTNU), Norway
}

\maketitle
\thispagestyle{empty}

\begin{abstract}

Recent studies have emphasized the potential of forehead-crease patterns as an alternative for face, iris, and periocular recognition, presenting contactless and convenient solutions, particularly in situations where faces are covered by surgical masks. However, collecting forehead data presents challenges, including cost and time constraints, as developing and optimizing forehead verification methods requires a substantial number of high-quality images. 
To tackle these challenges, the generation of synthetic biometric data has gained traction due to its ability to protect privacy while enabling effective training of deep learning-based biometric verification methods. In this paper, we present a new framework to synthesize forehead-crease image data while maintaining important features, such as uniqueness and realism. The proposed framework consists of two main modules: a Subject-Specific Generation Module (SSGM), based on an image-to-image Brownian Bridge Diffusion Model (BBDM), which learns a one-to-many mapping between image pairs to generate identity-aware synthetic forehead creases corresponding to real subjects, and a Subject-Agnostic Generation Module (SAGM), which samples new synthetic identities with assistance from the SSGM. We evaluate the diversity and realism of the generated forehead-crease images primarily using the Fréchet Inception Distance (FID) and the Structural Similarity Index Measure (SSIM). In addition, we assess the utility of synthetically generated forehead-crease images using a forehead-crease verification system (FHCVS). The results indicate an improvement in the verification accuracy of the FHCVS by utilizing synthetic data.

\end{abstract}

\begin{figure}[htp]
\centering
\includegraphics[width=0.47\textwidth]{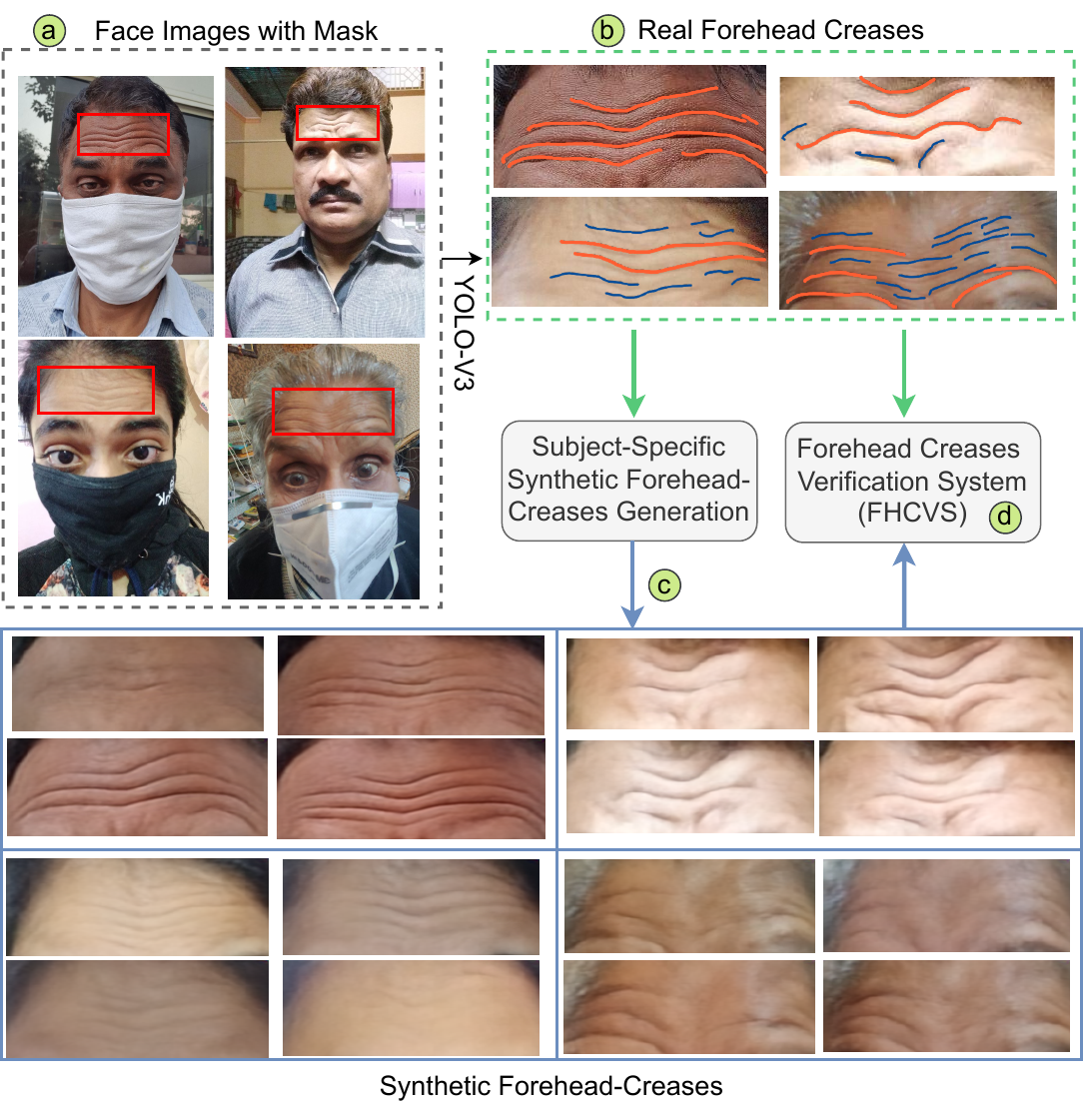}
\caption{(a) Capturing face images with forehead-crease using mobile devices, (b) Identifying the forehead region of interest (ROI) using YOLO-V3, (c) Generating subject-specific forehead creases images (d) Utilizing both real and synthetic forehead-crease images to train the forehead-creases verification system.}
\label{fig:teaser}
\end{figure}

\section{Introduction}
Contactless biometric recognition is a paradigm shift in identity verification, which utilizes advanced technologies to authenticate individuals without physical contact with sensors. The contactless approach includes various modalities such as face \cite{deng2019arcface}, iris \cite{daugman2009iris}, voice, and  gait \cite{thapar2019gait} recognition using touchless sensors. Recent advancements in the field of biometric recognition have introduced a new frontier: the utilization of forehead-creases patterns \cite{bharadwaj2022mobile} in masked (surgical) face scenarios. In circumstances where facial features are concealed, for example when wearing facial surgical masks, the characteristic patterns of creases on the forehead serve as valuable biometric identification markers. 
Forehead creases are biometric traits used to verify individuals based on the unique features of their forehead regions. 
Unlike other biometric traits, collecting data for forehead-creases recognition requires the user's consent because the creases are formed by facial muscle movements. This inherent requirement for user consent enhances the resistance to presentation attacks  of forehead-creases recognition compared with other crease-based biometric traits. The forehead has prominent features in the center and upper portions of the forehead, consisting of various layers of skin, muscle, and tissue. These muscles play a role in facial expressions such as raising the eyebrows and frowning to create unique forehead creases. These creases could be horizontal, vertical, wrinkled, or expression lines, as shown in Figure \ref{fig:teaser}. Forehead creases are unique to each individual and provide additional information for accurate and reliable user verification.

Since forehead creases biometrics were introduced recently \cite{bharadwaj2022mobile} \cite{sharma2024fh}, one of the primary challenges in developing robust forehead creases pattern recognition systems is the limited availability of real-world data. Unlike other facial features, data on forehead creases patterns are sparse, making it difficult to train accurate and reliable algorithms. To address this challenge, new methods are being applied to synthetic data generated by state-of-the-art generative models, such as GANs \cite{goodfellow2020generative}, VAE \cite{kingma2013auto}, and diffusion models \cite{ho2020denoising} can synthesize pseudo-biometric patterns that closely resemble real-world data. By harnessing the power of diffusion networks as generative models, we can make the effort to significantly expand the available dataset for training and testing forehead creases pattern recognition algorithms. Therefore, the importance of synthetic data in this context cannot be overstated. It not only mitigates the limitations imposed by data scarcity but also alleviates privacy concerns associated with collecting real-world biometric data. 

\begin{figure*}[!ht]
\centering
\includegraphics[width=1.0\textwidth]{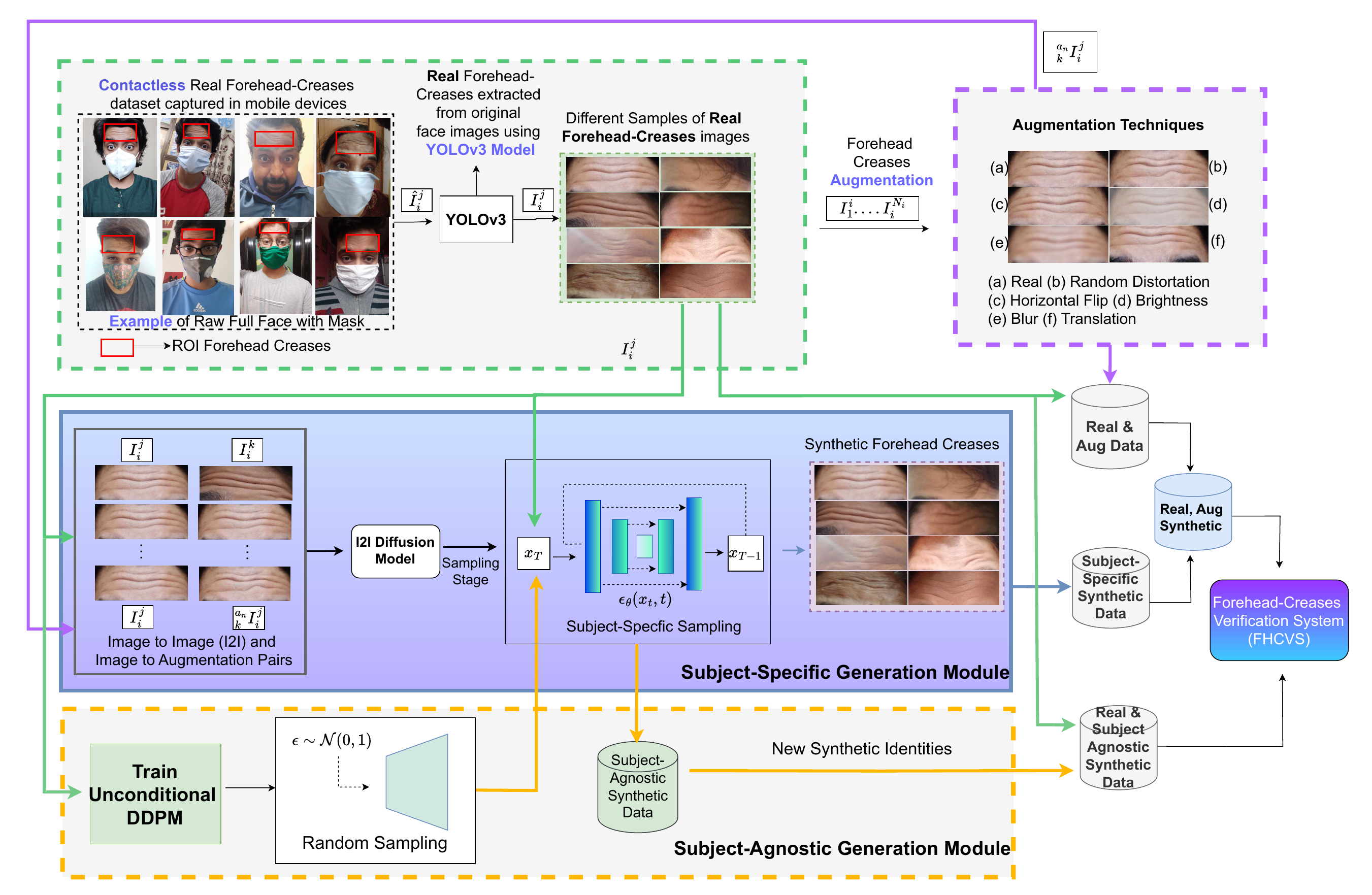}
\caption{Our proposed framework consists of two modules: Subject-Specific Generation Module (SSGM) and Subject-Agnostic Generation Module (SAGM). SSGM obtains image pairs for each subject from a forehead-crease image FH-V1 dataset \cite{bharadwaj2022mobile} and its augmentations to train an Image-to-Image translation-based diffusion model \cite{li2023bbdm}. $I_{i}^{j}$, i.e. the $j^{th}$ real pose of subject $i$, acts as a starting point $x_T$ for the pre-trained BBDM and generates subject-specific synthetic data. The SAGM samples random forehead creases from an unconditional DDPM \cite{ho2020denoising} trained on unlabeled forehead creases. These random samples act as an input to the sampler in SSGM, generating new identities. The synthetic datasets, merged with the real and augmented data are subsequently used as input to the forehead-creases verification backbone \cite{bharadwaj2022mobile}.}
\label{fig_main_figure}
\end{figure*}

\subsection{Our Contributions}
In this work, we focus on generating synthetic forehead-creases images with the sole objective of improving the performance of an existing forehead-creases recognition system trained using deep metric learning. The overall approach is illustrated in Figure \ref{fig_main_figure}  
We consider the following primary challenge: a lack of intra-subject diversity in the forehead-creases image dataset \cite{bharadwaj2022mobile}, along with an overall low number of subjects in the dataset. We address various challenges in the real-world use of forehead-creases images as a biometric, such as brightness and occlusion due to long hair, and aim to solve these challenges using subject-specific and subject-agnostic synthetic data. 
Our main contributions to this work are described as follows:
\begin{itemize} [leftmargin=*,noitemsep, topsep=0pt,parsep=0pt,partopsep=0pt]
\item	We propose a strategy to train an image-to-image diffusion bridge model using one-to-many mapping obtained from all possible image pairs of a labeled forehead-creases dataset. To increase the generative ability of the diffusion model, we train the diffusion bridge model with image-to-augmentation pairs to sample subject-specific diverse forehead poses while maintaining label consistency. \label{contribution:image-to-augmentation}	
\item	We sample random synthetic forehead-creases from an unconditional diffusion model and their variations from  subject-specific bridge diffusion sampler, resulting in new subject-agnostic synthetic identities.
\item Extensive experiments were conducted to evaluate the utility of the proposed synthetic forehead crease. To this extent, we incorporate various data combinations while training the forehead-crease verification model, ranging from real, synthetic, and augmented samples to mitigate the challenges in real test sets.
\item To enhance the reproducibility of our work, we make the code and synthetic forehead-creases biometric dataset readily available to the research community\footnote{ \href{https://github.com/abhishektandon/synthetic-forehead-creases}{https://github.com/abhishektandon/synthetic-forehead-creases}}.
\end{itemize}

The remainder of this paper is organized as follows. We discuss the existing literature on synthetic image generation and forehead  creases biometrics in section \ref{sec:related work}. The proposed approach is discussed in section \ref{sec:methodology}.  Section \ref{sec:db} presents the databases and evaluation protocols,  section \ref{sec:experiments} presents the experiments and results, and section \ref{sec:conc} concludes the paper. 
\section{Related Work} \label{sec:related work}
In this section, we first discuss the existing literature on forehead-crease biometric recognition, followed by the  synthetic generation of biometric characteristics. 

\subsection{Forehead Creases Recognition} \label{related work:forehead recognition}

Biometric recognition methods play a crucial role in modern security systems by providing reliable means of verification based on the unique physiological or behavioral traits of individuals. These methods leverage a diverse range of biometric modalities, including facial features, forehead creases \cite{bharadwaj2022mobile}, fingerprints\cite{maltoni2009handbook}, Knuckle-prints, iris patterns\cite{daugman2009iris}, voice characteristics, and gait dynamics\cite{thapar2019gait}, each of which offers distinct advantages and applications in various domains. However, forehead creases have been explored less in the field of biometrics. 

 Recently, researchers have focused on forehead-based segmentation and recognition methods. It started when COVID-19 became a major concern and wearing surgical face masks became a regular part of our everyday lives \cite{bharadwaj2022mobile}. Facial masks cover a large portion of the face, obscuring important facial features. This poses new challenges for face recognition systems. \cite{bharadwaj2022mobile} proposed to consider only the forehead-creases instead of complete face images as a new biometric trait for identification and verification of a person wearing a face mask. It employs deep metric learning for efficient forehead verification of smartphone selfies in various environments. The results show high performance with their proposed dataset of 247 subjects and 4,964 images introduced. Further, \cite{sharma2024fh} proposed FH-SSTNet, a 3D-CNN based architecture \cite{szegedy2015going} trained with triplet and ArcFace losses \cite{deng2019arcface} and performed extensive experiments on the same dataset \cite{bharadwaj2022mobile}.
 
\subsection{Synthetic Image Generation} 
Synthetic biometric data generation has become the popular in the biometric community due to the privacy preserving and availability of the large scale data can facilitate the training of complex deep learning model for verification. The motivation for evaluating synthetic forehead crease pattern generation for reliable verification was derived from the successful utilization of synthetic samples for other biometric characteristics.  
Our work is closely related to works such as  \cite{zhao2022bezierpalm} \cite{jin2024pce} \cite{kim2023dcface} which sampled synthetic biometric data to bridge the gap between the lack of training data and the performance of existing biometric systems, especially those based on lines, patterns, and creases such as palmprint, knuckle-creases, and fingerprints. 
RPG-Palm \cite{shen2023rpg} proposed a method to generate synthetic palmpprints and train a GAN-based unpaired image-to-image translation network to generate new identities. They used palmprint creases mathematically generated by manipulating the parameters of geometric Bezier curves as an external condition to the proposed generative network. They use an identity-aware loss to maintain the identity consistency of the generated poses. Their method improves upon the work of \cite{zhao2022bezierpalm} because they generate more realistic palmar creases as conditions for their proposed generative model. Using synthetic pre-training and fine-tuning on real data, they demonstrated improved results. PCE-Palm \cite{jin2024pce} further aims to bridge the gap between the realism of the generated palm prints by breaking the synthesis pipeline into two stages, each responsible for creases and textures. Their proposed palm-creases energy domain acts as a bridge between the Bezier curves and the real palmprint domain. Similar works, such as PrintsGAN \cite{engelsma2022printsgan} and \cite{wyzykowski2023synthetic} have been explored in fingerprint synthesis. However, all of these works rely heavily on GANs \cite{goodfellow2020generative}, which are notoriously difficult to train and suffer from mode collapse. There is a need to explore Diffusion Models \cite{ho2020denoising} to generate creases and pattern-based biometric data, which have shown great capability to generate diverse images using a process by gradually corrupting images with noise in the forward process and estimating the added noise using a neural network during the reverse process. For other biometrics, such as face, DCFace \cite{kim2023dcface} proposed a dual-conditioned diffusion model to generate a synthetic dataset by combining subject identity and external factors (i.e., style) of faces to train a diffusion model. 
Diffusion models have been extended to other vision tasks, such as text-to-image generation \cite{rombach2022high}, image editing \cite{mokady2023null}, inpainting \cite{avrahami2023blended}, and image-to-image translation \cite{su2022dual}. BBDM \cite{li2023bbdm} is one the first image-to-image translation networks to introduce a bridge diffusion model between image pairs as endpoints without relying on the conditioning mechanism of latent diffusion models \cite{rombach2022high}. It attempts to directly learn the distribution of the target domain, given the input image distribution in the reverse diffusion process. Image-to-image translation is relevant in our work; therefore, we utilized this network for our task. Generating forehead-creases images is an unexplored area, and we aim to explore the same in our work.

\section{Proposed Approach } \label{sec:methodology}
The proposed approach of the forehead-creases image generation framework is divided into two modules: a Subject-Specific Generation Module , which is based on an image-to-image translation network BBDM \cite{li2023bbdm} and a Subject-Agnostic Generation Module  based on DDPM \cite{ho2020denoising}.  A block diagram of the proposed approach is shown in Figure \ref{fig_main_figure} that has four different functional blocks: (1) ROI extraction (2) Subject-Specific Generation Module (3) Subject-Agnostic generation module and (4) Forehead-creases based user verification Network.

\subsection{ROI extraction} \label{roi}
Given a facial image of subjects with forehead creases, we segment the Region of Interest (ROI), i.e., the coordinates of the forehead creases using a finetuned YOLOv3 \cite{redmon2018yolov3} to obtain a forehead-creases image dataset containing $D$ subjects labeled as $\{1, 2, \ldots, i, \ldots, D\}$. The segmentation performance is the same as described in \cite{bharadwaj2022mobile}. Each subject has varying poses, resulting in $N$ different images, and the $j^{th}$ pose is denoted as $I_{i}^{j}$. Each such pose is augmented using 6 augmentation techniques: translation, random distortion, horizontal flip, brightness, blur, and occlusion. Each augmented image can be expressed as ${ }_k^{a_n} I_i^j$ denoting the $k$-th image of augmentation $a_n$; $n \in \{1, 2, \ldots, 6\}$. Example of the ROI images and the corresponding augmented images are as shown in the  Figure  \ref{fig_main_figure}. 

\subsection{Subject-Specific Synthetic Forehead-Creases Generation}
\label{subsec:ssgm}
Subject-Specific Generation Module (SSGM) as shown in Figure \ref{fig_main_figure}, aims to synthesize diverse intra-subject forehead-creases images conditioned on a given pose $I_{i}^{j}$. 
It is known that high intra-class diversity is desired in recognition tasks. 
However, subject-specific generated images should not be too diverse and label consistency should also be maintained. An image-to-image translation model is the perfect choice to solve such a problem. 
We utilize the Brownian Bridge Diffusion model (BBDM) \cite{li2023bbdm} network for the generation task. Note that DDPM \cite{ho2020denoising} follows a stochastic trajectory via noise injection, which cannot be used to obtain a mapping between two images directly unless it is conditioned using attention mechanism \cite{rombach2022high}, it rather maps the image from the image domain to isotropic Gaussian noise. BBDM re-defines the DDPM trajectory to make the diffusion process work for image pairs to estimate a target distribution, rather than a single input image distribution. Mathematically, it defines the forward diffusion trajectory such that $t=0$ results in the initial image $x_{0}$, while $t = T$ results in the target image $y$ at $x_{T}$, hence tying down the trajectory between two endpoints. Thus, the forward process is formulated as:
\begin{align}
    q(x_{t} | x_{0}, x_{T}) = \mathcal{N}\big(x_{t}; \big(1 - m_t \big) x_{0} + m_{t} x_{T}, \delta_{t}I\big)
    \label{equation:bbdm_forward}
\end{align} 
Here, $m_{t}$ = $\frac{t}{T}$ and the variance $\delta_{t}$ is defined as:
\begin{align}
    \delta_{t} = 2m_{t}\big(1 - m_{t} \big) \label{eq_var}
\end{align}
During the reverse process, it aims to predict the joint distribution of $x_{t-1}$ given $x_{t}$, and a target image $y$ i.e. $p_{\theta}(x_t|x_{t-1}, y)$ which is used as a starting point, instead of noise.  This is done using a neural network, similar to that of DDPM, thus the BBDM training objective can be simplified as:
\begin{align}
    \mathbb{E}_{{x}_0, {y}, {\epsilon}}\big[|| \epsilon_{\theta}(x_{t}, t) - (\sqrt{\delta_{t}}\epsilon + m_t(y - x_{0})) ||^2\big]
\end{align}
While sampling, the target image is set only as the initial starting point to produce diverse image samples while preserving class identity. 
The variance term $\delta_{t}$ can also be multiplied by a factor of $s$ to increase the diversity of sampled images. 
\begin{figure}[!ht]
\begin{center}
    \includegraphics[width=1.0\linewidth]{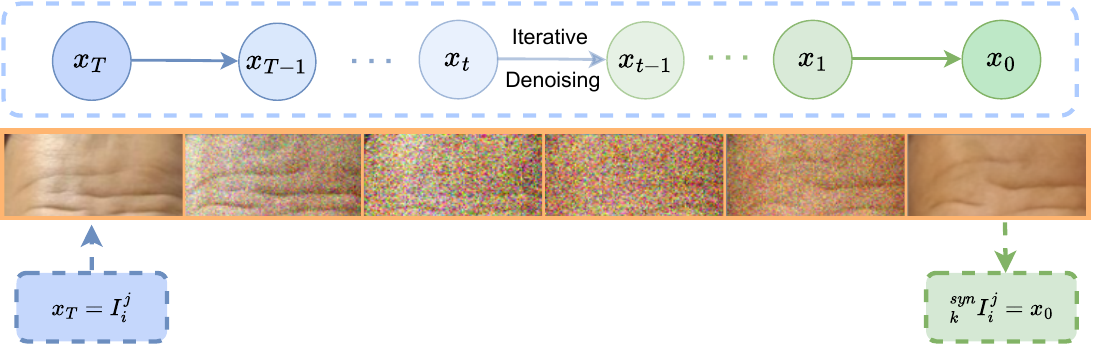}
\end{center}
   \caption{Sampling Subject-Specific Synthetic Data: It sets a pose $I_{i}^{j}$ as a starting point to the BBDM \cite{li2023bbdm} sampler, trained on forehead-creases image pairs. After gradual denoising, the final obtained synthetic image ${}^{syn}_{k} I_{i}^{j}$ is added to the training data of the corresponding identity $i$.}
\label{fig_testing_bbdm}
\end{figure}
SSGM utilizes real forehead-creases data to obtain a rich set of image pairs. These image pairs are further used as training data for the image-to-image (I2I) translation-based Brownian Bridge Diffusion Model (BBDM) \cite{li2023bbdm}. For a given labeled real forehead-creases dataset with $D$ subjects containing $N$ images each, we take all the permutations to obtain $\Perm{N}{2}$ image-pairs for each subject, resulting in both forward $(I_{i}^{j} \rightarrow I_{i}^{k})$ and backward pairs $(I_{i}^{k} \rightarrow I_{i}^{j})$, hence training the I2I model with one-to-many image mappings. 
The image at time step $t$ can be obtained directly from $x_0$ from equation \ref{equation:bbdm_forward} as:
\begin{align}
    x_t = (1 - m_t)x_0 + m_tx_T + \sqrt\delta_t\epsilon
\end{align}
To further enhance generation quality and diversity, we randomly select image-to-augmentation $(I_{i}^{j} \rightarrow { }_k^{a_n} I_i^j)$ forward pairs to the training paired images database. Along with the $\Perm{N}{2}$ pairs, each pose is mapped to its corresponding augmentation; particularly translation, random distortions, blur, and brightness. To reduce the training time complexity, 100 such image-to-augmentations pairs are sampled randomly for a given subject. This facilitates the model to capture high variations in image mappings and generate diversified samples on inference. This is evident from Table \ref{table_fid_diversity}, that our image-to-augmentation training strategy improves generation quality as it achieves the lowest FID score and maximum diversity score. 

\begin{figure}[htp]
\centering
\includegraphics[width=0.45\textwidth]{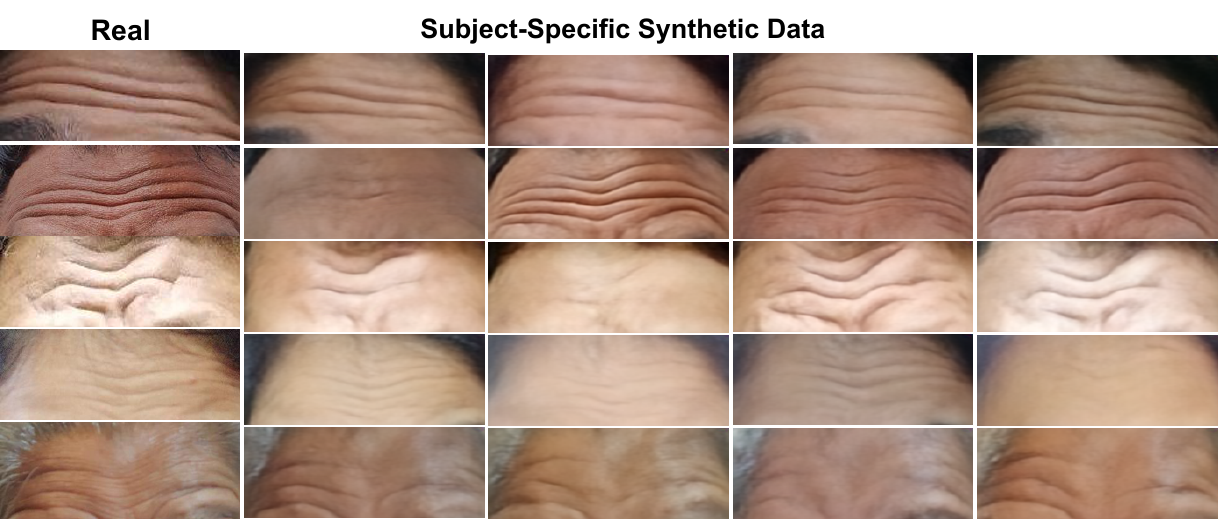}
\caption{Subject-Specific (SS) Samples: FH-V1 Real images (first column) of different subjects (IDs: 70, 195, 203, 207, and 239) and their synthetic counterparts (subsequent columns), highlighting the diversity in synthetic image generation.}
\label{fig_synthetic_samples}
\end{figure}

During sampling, a subject pose $I_{i}^{j}$ is used as input (or starting point $x_T$ for the diffusion sampler, depicted in Figure \ref{fig_testing_bbdm}) to the trained I2I diffusion model to generate diverse subject-specific synthetic data as shown in Figure \ref{fig_synthetic_samples}, denoted as ${}_{k}^{syn} I^{j}_{i}$.
We train the BBDM network for 200 epochs using Adam optimizer and a learning rate of $10^{-4}$. The diffusion time step $T$ is set to 1000 while training and 200 while sampling. We train our model on a single NVIDIA Tesla V100-SXM2 GPU with 16 GB V-RAM.
We set maximum variance $s = 1$, thereby sampling at a low variance as per equation \ref{eq_var} to maintain label consistency. 

\subsection{Subject-Agnostic Synthetic Forehead-Creases Generation} \label{sagm}

The goal of subject-agnostic synthetic forehead-creases generation is to generate unique synthetic IDs that can facilitate privacy-preserving forehead-creases and enable large data for training. To establish the efficacy of our generation strategy, we sampled random identities, agnostic to real subjects, and trained the verification system on these identities.  

\begin{figure}[htp]
\centering
\includegraphics[width=0.45\textwidth]{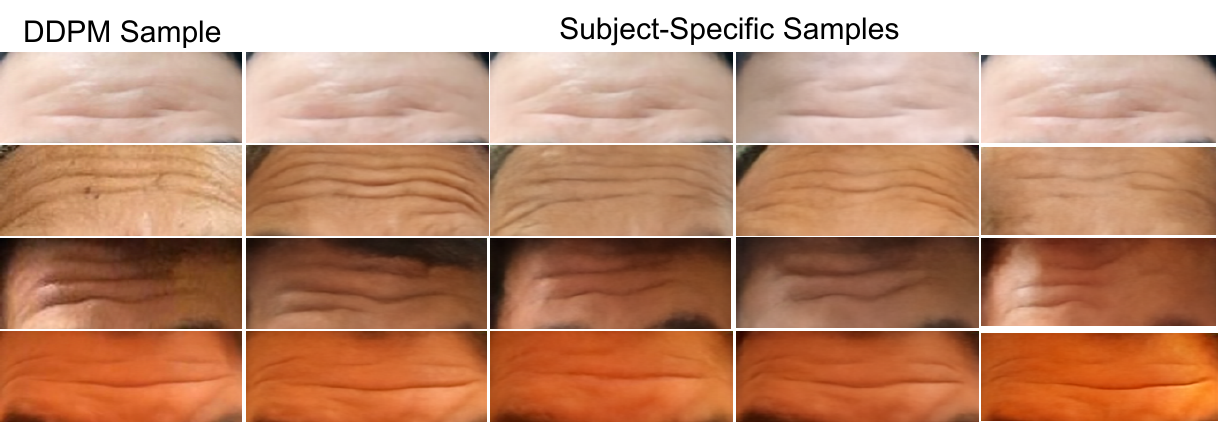}
\caption{Subject-Agnostic (SA) Synthetic Data: Random sample from DDPM (left) and corresponding subject-specific samples (rest).}
\label{SA_samples}
\end{figure}

The Subject-Agnostic Generation Module (SAGM) takes unlabeled real forehead-crease images to train an unconditional DDPM \cite{ho2020denoising}. After training, the pre-trained unconditional model takes Gaussian noise $\epsilon \sim \mathcal{N}(0, 1)$ as input, and iteratively denoises it to generate random samples. These random samples act as input to the subject-specific generator of SSGM to sample different poses, resulting in 247 new synthetic identities, agnostic to the real subjects. These new synthetic identities are merged with the real training set, resulting in a larger dataset. The generated samples are presented in Figure \ref{SA_samples}.

\subsection{Forehead-creases User Verification Network} \label{subsec:recognition network}

To benchmark the effectiveness of the proposed synthetic generation of forehead-creases, we consider the forehead-creases user verification system from \cite{bharadwaj2022mobile} owing to its effectiveness and robust performance.  We refer to this as Forehead-creases Verification System (FHCVS). The FHCVS uses ResNet-18 \cite{he2016deep} to extract features from forehead-creases images of size $224\times224$. ResNet-18 captures hierarchical representations of features, facilitating discrimination. The Position Attention Module (PAM) \cite{fu2019dual} preserves spatial information, while the ECA attention module \cite{wang2020eca} refines features. Fully connected layers produce a 512-dimensional embedding for verification, enhancing performance. The network is trained using AdaFace \cite{kim2022adaface} margin loss with $m$ = 0.4, $h$ = 0.33 and $s$ = 64.0, followed by focal loss \cite{lin2017focal} with $\gamma$ = 2.
Given the enrolled and probe images, the comparison scores were computed using  Euclidean distance.

\section{Dataset Description and Evaluation protocol}
\label{sec:db}

\subsection{Real Datasets} The FH-V1 dataset \cite{bharadwaj2022mobile}  is publicly available and comprises 247 data subjects, whose forehead-creases were captured using a smartphone. It was collected across two sessions with a gap of one day \cite{bharadwaj2022mobile}, each consisting of two poses.  In Pose 1, images were captured at a fixed distance between the camera and phone. In Pose 2, the users were instructed to minimize the distance between the camera and phone. The poses were captured under diverse conditions, including variations in background, brightness, occlusion, and angles. Each subject had 10-20 samples collected during both sessions. Additionally, more data was collected using a similar approach, thereby increasing the subjects by 295 subjects, resulting in the FH-V2 dataset with a total of $542$ subjects. We call the additional 295 subjects as \textbf{(FH-V2)${}^{'}$} dataset. This data is also made publicly available for the research community.

\subsection{Synthetic Dataset:}
In this work, we have proposed a synthetic forehead-creases generation using the Brownian Bridge Diffusion model (BBDM) \cite{li2023bbdm} model (see Section \ref{sec:methodology}) to generate (a) a subject-specific (SS) and subject-agnostic (SA) synthetic forehead-creases database.  As discussed in Section \ref{subsec:ssgm}, we first trained the I2I generative network with $\sum_{i}^{D} \Comb{N_i}{2}$ image pairs, resulting in a total of $11050$ pairs ($\sim$ 45 pairs per subject). Here, $D$ is the total number of real identities and $N_i$ is the number of poses for the $i^{th}$ identity in the \textbf{FH-V1 dataset}, that is, $247$. Further, we increased the number of pairs to $\sum_{i}^{D} \Perm{N_i}{2} = 22100$ ($\sim$ 90 pairs per subject). After adding 100 random augmentation pairs, the resulting number increases to $\sum_{i}^{D} \Perm{N_i}{2} + 100 \times 247 = 46800$ pairs. Therefore, our training strategy results in three I2I models: $Comb,\ Permute,\ and\ PermuteAug$. \label{synthetic-datasets}

\subsubsection{Subject-Specific (SS) Synthetic dataset}
We introduce subject-specific synthetic datasets that include variations in the amount and intensity of muscle contraction, poses, and image intensity. For comparison with a GAN-based I2I model, we also train BiCycleGAN \cite{zhu2017toward} under the same setting used to train BBDM \cite{li2023bbdm} as mentioned in Section \ref{synthetic-datasets} i.e. forward and backward image pairs (Permute), as well as image-to-augmentation pairs (PermuteAug).
Furthermore, for each image in the FH-V1 training dataset, we sample five synthetic variations using the pre-trained I2I models resulting in a total of $5 \times 2462 = 12310$ images per dataset.

\subsubsection{Subject-Agnostic (SA) Synthetic dataset}

For subject-agnostic datasets, we added a total of 10 poses generated using the best-performing $PermuteAug$ model (as per Table \ref{table_fid_diversity}). A randomly sampled forehead-creases (total $247$) from an unconditional DDPM was used as a conditional starting point for the FH-BBDM network, resulting in a total of $247 \times 11 = 2717$ images per dataset (including original sample). The resulting datasets are named $SS$ and $SA$ (Subject-Specific and Subject-Agnostic, respectively), followed by the I2I model name used to generate the poses. $BG$ denotes BiCycleGAN \cite{zhu2017toward}. Table \ref{table_synthetic_data_info} lists the statistics for the synthetic dataset generated in this work. 
\begin{table}[htbp]
    \centering  
    \label{tab:my-table2}
    \begin{tabular}{ccccccc}
    \hline
    \textbf{Synthetic Dataset} & \textbf{Subjects} & \textbf{Images} \\ \hline
    SS-Comb & 247 & 12310  \\ 
    SS-Permute & 247 & 12310  \\ 
    SS-PermuteAug & 247
 & 12310  \\ \hline
    SS-BG-Permute & 247 & 12310  \\ 
    SS-BG-PermuteAug & 247
 & 12310  \\ \hline
 SA-PermuteAug & 247 & 2717  \\ \hline
    \end{tabular}
    \caption{Statistics of Synthetic Datasets generated in this work.}
    \label{table_synthetic_data_info} 
\end{table}
 
\subsection{Performance Evaluation Protocol}
To evaluate the utility of the proposed synthetic forehead-creases generation, the benchmark performance of the FHCVS (Forehead-creases Verification System) \cite{bharadwaj2022mobile} on forehead-creases datasets was evaluated using three different experimental protocols. 

\begin{table*}[htbp]
    \centering
    \label{tab:my-table3}
    \begin{tabular}{lccccc}
    \hline
    \multirow{2}{*}{\textbf{Experiments}} & \multirow{2}{*}{\textbf{Database}} & \multirow{2}{*}{\textbf{EER (\%) \(\downarrow\)}} & \multicolumn{2}{c}{\textbf{TMR (\%) @ FMR (\%) =}} \\ \cline{4-5} 
     &  &  & \textbf{0.1 (\%) \(\uparrow\)} & \textbf{0.01 (\%) \(\uparrow\)} \\ \hline
    \multirow{3}{*}{\textbf{Experiment 1}} & FH-V1         & 12.39  & 40.19 &  21.97 \\
    & FH-V1 with Translation   & 10.23  & 55.26 & 40.41 \\
    & FH-V1 with all Aug &  11.25 & 55.20 &  40.71   \\ \hline
    
    \multirow{5}{*}{\textbf{Experiment 2}}& SS-Comb & 11.83  & 49.94 & 34.42 \\
    & SS-Permute &  10.37  & 56.06 & 40.57  \\
    & \textbf{SS-PermuteAug} & \textbf{9.38} &  \textbf{60.32} & \textbf{45.68} \\

    & FH-V1 with all Aug + SS-Permute   & 10.46 & 57.69 & 41.95  \\ 
    & FH-V1 with all Aug + SS-PermuteAug & 11.20 &  56.13   & 41.78 \\    \hline

    
     
    \multirow{5}{*}{\textbf{Experiment 3}} & SS-BG-Permute &  14.92  & 25.05 & 10.33  \\
    & SS-BG-PermuteAug & 17.67 & 22.93 & 9.68 \\
    & FHV1 with all Aug + SS-BG-Permute & 11.06 &  57.09 & 43.81 \\
    & FHV1 with all Aug + SS-BG-PermuteAug &  11.78 & 53.97  & 39.62  \\
    
    \hline
    \end{tabular}
    \caption{Quantitative results on FH-V1 and Subject-Specific (SS) synthetic datasets.}
    \label{table:final table}
\end{table*}

\begin{itemize} [leftmargin=*,noitemsep, topsep=0pt,parsep=0pt,partopsep=0pt]
\item \textbf{Experiment 1:} In this experiment, the performance of the FHCVS on the real forehead-creases dataset FH-V1. We present three different experiments: (a) training FHCVS without data augmentation, (b)  training FHCVS with data augmentation (translation), and (c) training FHCVS with data augmentation (brightness, occlusion, blur, random distortion, and translation). The trained network is used as a feature extractor to obtain embeddings for both the enrolment and probe samples from the (FH-V2)$^{'}$ dataset to evaluate the performance.

\item \textbf{Experiment 2:}  In this experiment, a subject-specific synthetic dataset was used both independently and with real data to train the FHCVS. However, the testing set corresponds to the real data subjects. The trained network is used as a feature extractor to obtain embeddings for both the enrolment and probe samples from the (FH-V2)$^{'}$ dataset to evaluate the performance.

\item \textbf{Experiment 3:} 
This experiment is similar to Experiment 2, except that the subject-specific synthetic data is sampled using BiCycleGAN \cite{zhu2017toward}, instead of BBDM \cite{li2023bbdm}, and we evaluate the performance of FHCVS on (FH-V2)$^{'}$

\end{itemize}

\section{Experimental Results}
\label{sec:experiments}

In this section, we present the qualitative and quantitative performances of the proposed synthetic forehead-creases biometric data. A qualitative analysis of the generation quality is presented using the Fréchet Inception Distance (FID) \cite{heusel2017gans}, intra-subject diversity as proposed in \cite{batzolis2021conditional} and  SSIM \cite{wang2004image}. The quantitative results are presented using verification performance, which is quantified using the Equal Error Rate (EER), False Match Rate (FMR), and True Acceptance Rate (TAR) which is equal to 100 minus False Non-Match Rate (FNMR), TMR (\%), and FMR (\%) at 0.1\% and 0.01\%. The EER was computed at the point where the FMR equals to FNMR.

\subsection{Qualitative Evaluation of synthetic Forehead-creases Generation} 
\begin{figure}[htp]
\centering
\includegraphics[width=0.45\textwidth]{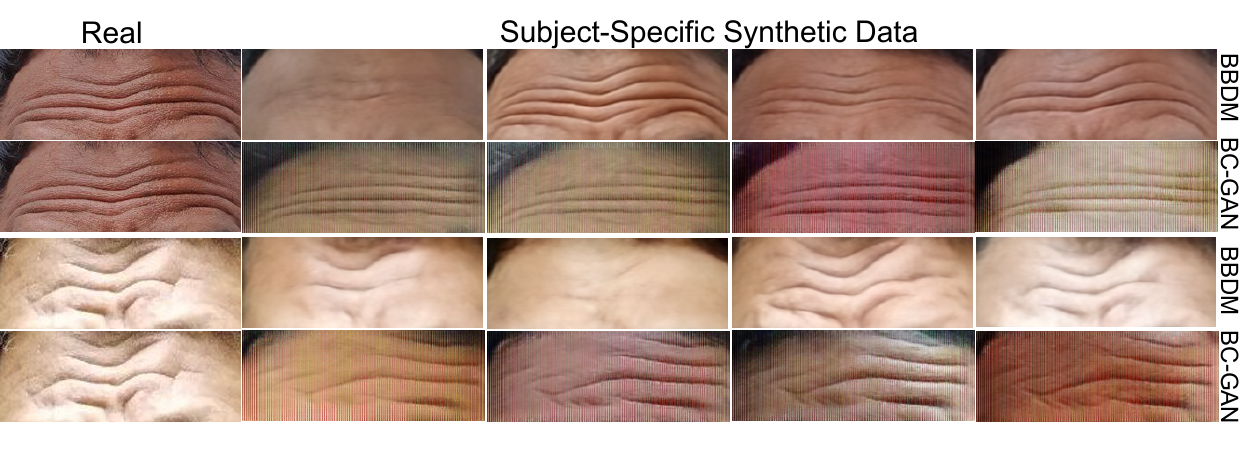}
\caption{Comparison: Real images from FH-V1 of two different subjects (IDs: 195, and 203) in the first column. $1^{st}$ and $3^{rd}$ rows display synthetic images generated using BBDM, while the $2^{nd}$ and $4^{th}$ rows are images generated by BicycleGAN. However, a few generated images using BBDM also lack creases ($2^{nd}$ image in $1^{st}$ row), which suggests the sampler's failure in some cases.}
\label{fig:bicyclegan}
\end{figure}

The proposed model is evaluated based on the FID score, intraclass diversity of synthesized images, and average SSIM score.  In case of subject-specific data, the FID compares the closeness of the distribution of the generated forehead-creases pose variations to the ground truth (real forehead-creases poses acting as target $y$ while training the I2I model). The SSIM score is computed between a real subject pose and its corresponding pose-specific synthetic variation and averaged over all subjects in the real dataset. For subject-agnostic data, FID is computed between the sampled and real FH-V1 datasets and SSIM is between randomly sampled forehead-creases images and their subsequent variations. The performance metrics are presented in Table \ref{table_fid_diversity} which shows the good SSIM values for both SS-PermuteAug and SA-PermuteAug indicating the high quality of the generated images. For SS-BicycleGAN (BG) data, the FID score is high; the generated samples have poor perceptual quality, although the diversity of the samples is higher. The SSIM score is too low, suggesting the lack of label consistency amongst generated samples, also evident from Figure \ref{fig:bicyclegan}.

\begin{table}[htbp]
    \centering
    
    \label{tab:my-table2}
    \begin{tabular}{ccccccc}
    \hline
    \textbf{Synthetic Dataset} & \textbf{FID}  \(\downarrow\) & \textbf{Diversity} \(\uparrow\) & \textbf{SSIM} \(\uparrow\) \\ \hline
    SS-Comb& 63.253 & 23.646 & 0.723  \\ 
    SS-Permute& 53.257 & 22.454 & 0.695  \\ 
 \textbf{SS-PermuteAug }& \textbf{31.940}
 & 24.709 & \textbf{0.847} \\ 
    \hline
    SA-PermuteAug & 55.999 & 11.236 & 0.834 \\ \hline
    SS-BG-Permute & 256.752 & 35.517 & 0.199  \\ 
    SS-BG-PermuteAug & 232.417 & \textbf{37.457} & 0.147  \\ \hline
    
    \end{tabular}
    \caption{Qualitative evaluation of the forehead creases biometric Synthetic Data.}
    \label{table_fid_diversity} 
\end{table}

\subsection{Cross-Database Verification Results}
\begin{figure}[htbp]
\centering
\includegraphics[width=0.4\textwidth]{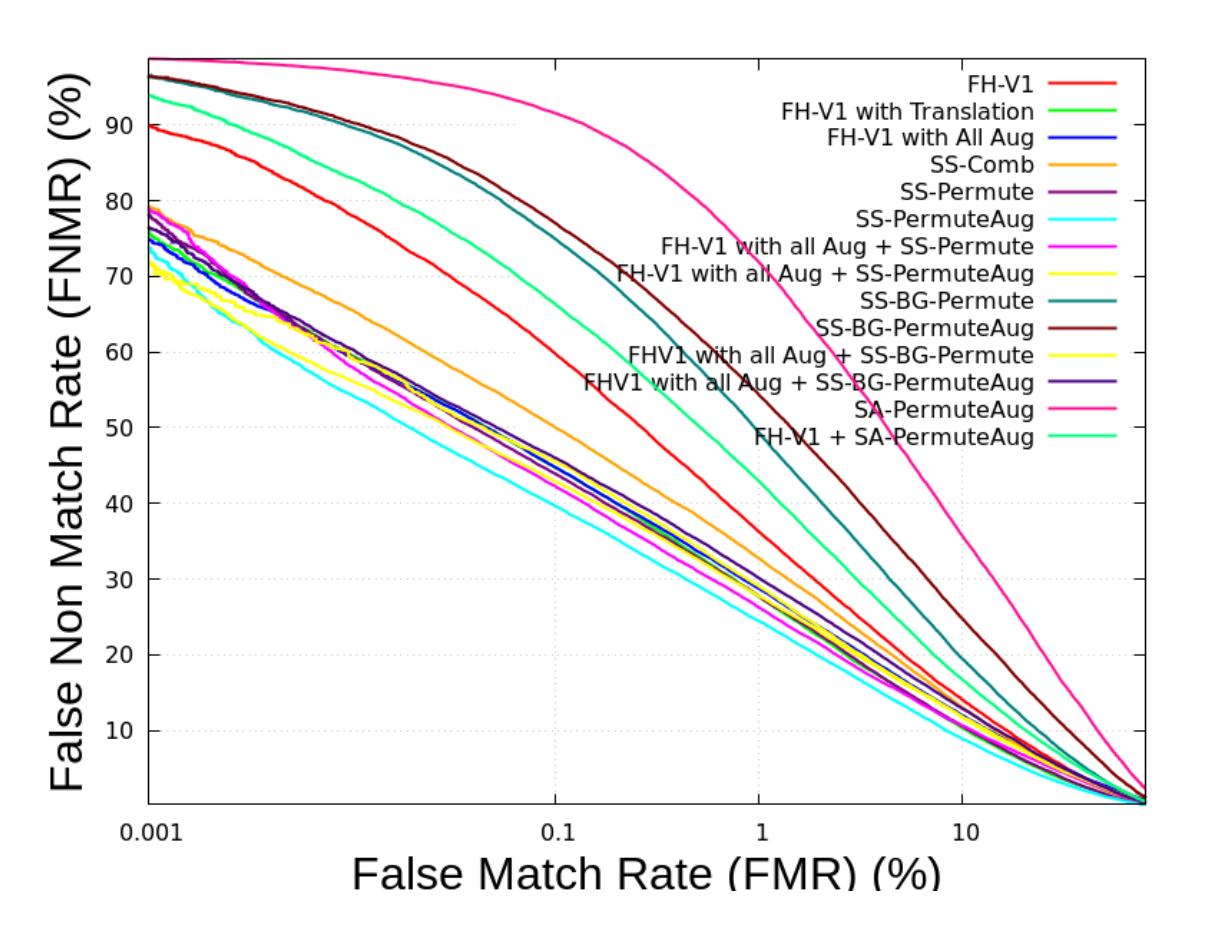}
\caption{DET curve for comparative analysis using different experiments with Forehead-creases real dataset, with augmentation, SS and SA synthetic datasets. The
x-axis is shown in log scale.}
\label{fig:table_5_roc}
\end{figure}

We conducted cross-database verification experiments as presented in Table \ref{table:final table}. We trained the FHCVS on real (FH-V1) and subject-specific synthetic databases (Table \ref{table_synthetic_data_info}) and used (FH-V2)${}^{'}$, i.e. unseen subjects, to evaluate the performance.
The FHCVS achieved an EER of $12.39\%$ with TMR and FMR values of $40.19\%$ and $21.97\%$, respectively, when trained with FH-V1 data without any augmentations. We studied the impact of applying augmentation techniques and subject-specific synthetic data on the performance of FHCVS. Experiment using augmentation techniques achieves an EER of $11.25\%$ with TMR and FMR values of $55.20\%$ and $40.71\%$, while using only translation is better (EER $10.23\%$, TMR $55.26\%$ and FMR $40.41\%$).
The experiments conducted on purely subject-specific synthetic datasets (Table \ref{table:final table}, first three rows in Experiment 2) significantly outperformed the model trained solely on real images (i.e. FH-V1 database). Our experiments also suggest that training the I2I diffusion model with both forward and backward image pairs (EER $10.37$\%, TMR $56.06$\%, and FMR $40.57$\%) enhances the verification performance of FHCVS compared to only forward pairs (EER $11.83\%$, TMR, and FMR of $49.94\%$ and $34.42\%$). Moreover, adding random image-to-augmentation pairs is even more effective, i.e., an EER of $9.38\%$, with TMR and FMR values of $60.32\%$ and $45.68\%$ respectively, which is the best-performing model, outperforming models trained on a combination of subject-specific datasets and real augmented data. Overall results are compiled in a combined DET plot in Figure. 
\ref{fig:table_5_roc}. The subject-specific BicycleGAN (BG) data is largely unsuitable for the recognition task, which is apparent from Experiment 3, as it achieves a higher EER compared to data sampled from BBDM \cite{li2023bbdm}. This also establishes the superiority of the chosen diffusion model instead of a GAN.


\subsubsection{Utility of Subject-Agnostic Synthetic Data}
To test the efficacy of our framework, we use the sampled subject-agnostic synthetic dataset (SA-PermuteAug) to train the FHCVS and report the cross-database verification results. The trained FHCVS was subsequently employed as the pre-trained network to extract features from the probe samples of the (FH-V2)${}^{'}$ dataset for the computation of the verification performance. This experiment aimed to investigate the verification performance of FHCVS when trained solely with synthetic identities. As a preliminary step, we first make sure that none of the subject-agnostic synthetic images match the existing real identities. We take the best-performing FHCVS model (Table \ref{table:final table}) to obtain the embeddings of the subject-agnostic images and match them with real image embeddings in the train set. We eliminate any synthetic identity that matches a real image with a certain threshold. 
Training FHCVS with this data, however, achieved a high EER of 22.37\% with TMR and FMR values of $8.44\%$ and $2.88\%$, respectively. Merging this data with real data instead, improves the EER to $13.76\%$, with TMR and FMR values of $33.62\%$ and $17.04\%$, but higher than that of training the model on FH-V1 alone, also reflected in Figure \ref{fig:table_5_roc}. The reason for such a poor performance can be attributed to the low diversity of the sampled data (Table \ref{table_fid_diversity}).

\section{Conclusion}
\label{sec:conc}

Synthetic biometric sample generation is essential for augmenting data to train large models while preserving privacy. We present a synthetic forehead-creases generation framework using the Brownian Bridge Diffusion model (BBDM). Both Subject-Specific and Subject-Agnostic forehead crease biometric samples were generated to address the challenges of small-scale datasets. Extensive experiments were performed to benchmark both qualitatively and quantitatively the utility of syntactic forehead crease biometric samples. The qualitative performance of our proposed approach was evaluated based on FID scores for distribution closeness and SSIM scores for the similarity between real and synthetic forehead crease variations. Quantitative experiments were performed on different combinations of real and synthetic forehead crease datasets. The benchmarking results with state-of-the-art  FHCVS achieved a cross-database EER of $9.38\%$  when trained on subject-specific synthetic data, demonstrating the effectiveness of our approach.

{\small
\bibliographystyle{ieee}
\bibliography{ref}
}

\end{document}